%% file: paper_final.tex
\documentclass[10pt,twocolumn,letterpaper]{article}

\usepackage{cvpr}
\usepackage{times}
\usepackage{epsfig}
\usepackage{graphicx}
\usepackage{amsmath}
\usepackage{amssymb}
\usepackage{multirow}
\usepackage{amssymb}
\usepackage[dvipsnames]{xcolor}
\usepackage{soul}
\usepackage[breaklinks=true,bookmarks=false]{hyperref}

\cvprfinalcopy 

\def\cvprPaperID{} 

\ifcvprfinal\fi
\begin{document}

\title{Focus on defocus: bridging the synthetic to real domain gap for depth estimation}

\author{Maxim Maximov \\
    Technical University Munich\\
    \and
    Kevin Galim\\
    Technical University Munich\\
    \and
    Laura Leal-Taix\'{e}\\
    Technical University Munich\\
}

\maketitle
\thispagestyle{empty}

\begin{abstract}
   Data-driven depth estimation methods struggle with the generalization outside their training scenes due to the immense variability of the real-world scenes. This problem can be partially addressed by utilising synthetically generated images, but closing the synthetic-real domain gap is far from trivial.  
   In this paper, we tackle this issue by using domain invariant defocus blur as direct supervision.
   We leverage defocus cues by using a permutation invariant convolutional neural network that encourages the network to learn from the differences between images with a different point of focus. 
   Our proposed network uses the defocus map as an intermediate supervisory signal. 
  We are able to train our model completely on synthetic data and directly apply it to a wide range of real-world images. 
   We evaluate our model on synthetic and real datasets, showing compelling generalization results and state-of-the-art depth prediction.
   The dataset and code are available at \url{https://github.com/dvl-tum/defocus-net}.
\end{abstract}

\section{Introduction}

In recent years, we have seen an increase in the number of smartphone photography users, bringing the need for image editing tools to a wider audience. 
Most of these tools are still limited to color adjustments and simple image transformations. More advanced post-capture changes such as focus and depth-of-field adjustments are not commonly available due to the need for depth maps of the captured scene.
While there exist specialized hardware solutions to compute depth, \eg, light-field cameras~\cite{DBLP:conf/cvpr/JeonPCPBTK15}, nowadays, data-driven machine learning makes it possible to tackle the problem from the software side, predicting depth maps~\cite{Guo18, DBLP:conf/cvpr/FuGWBT18,Yin_2019_ICCV} from a single image. 
However, monocular depth estimation methods do not generalize well to unseen data/scenes, \eg, different viewing angles, geometry and objects types. 
They heavily rely on perspective, size, texture and shading cues to measure distance. Those cues are dependent on the type of scene and objects, texture and illumination, which easily leads to overfitting to those memorized settings~\cite{Mancini17}.
Other works on depth estimation show better generalization by relying on comparison-based depth cues, such as depth from motion \cite{DBLP:conf/cvpr/UmmenhoferZUMID17, DBLP:conf/cvpr/MahjourianWA18} or stereo \cite{DBLP:conf/cvpr/ZhangPYT19, DBLP:conf/cvpr/SmolyanskiyKB18}.

\begin{figure}[t]
\begin{center}
   \includegraphics[width=0.48\textwidth, trim={0.0cm 1.5cm 0cm 0.0cm},clip]{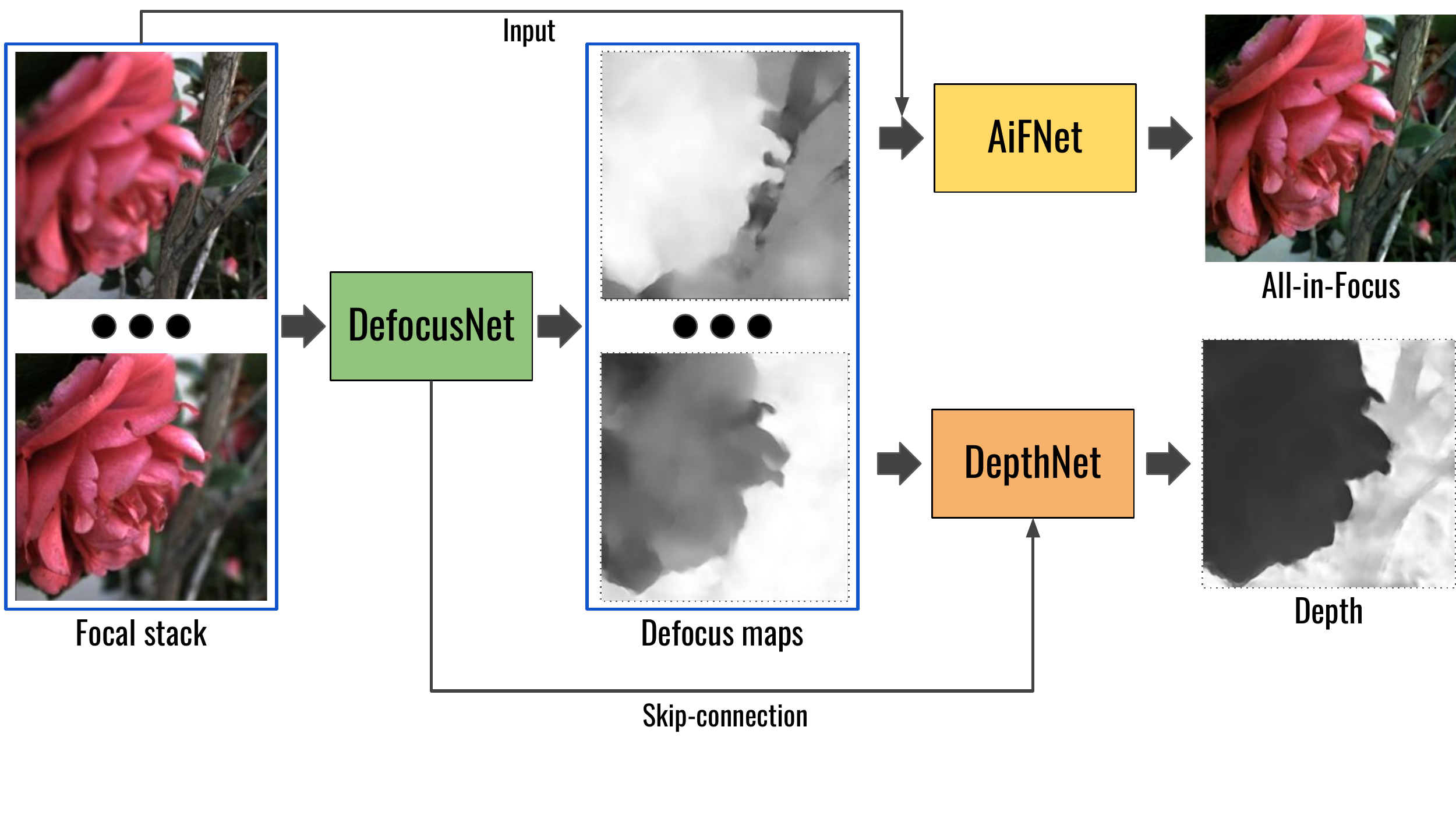}
\end{center}
   \caption{The pipeline of our approach. Our proposed end-to-end learned model combines depth and all-in-focus estimation from a focal stack using intermediate defocus map estimation and permutation-invariant networks, leading to a better generalization from a synthetic training to real photos.}
\label{fig:long}
\label{fig:cover}
\end{figure}

An under-explored cue for depth estimation is {\it defocus}, given that an object’s depth dictates how sharp it will appear in the image.
Depth-from-focus (or defocus) is defined as the task of obtaining the depth of a scene from a focal stack, \ie, a set of images taken by the same camera but focused on different focal planes. 
Analytical approaches \cite{moeller2015variational, Suwajanakorn15} compute depth based on the sharpness of each pixel. 
Such approaches are time-consuming, and perform especially poorly for texture-less objects. Recent deep learning approaches address these challenges in a data-driven way by learning to directly regress depth from a focal stack \cite{Hazirbas18}. 
Their main drawback is that they do not consider the underlying image formation process, therefore, such methods are also prone to overfitting to the specific training conditions.

Another challenge towards achieving generalization is the lack of high-quality and diverse training sets.
Collecting focal stack images with registered depth maps is an extremely time-consuming task, not to mention the imperfect depth ground truth data obtained from hardware solutions like time-of-flight sensors \cite{Hazirbas18}. 
One can rely on synthetic data as used in inverse-graphics tasks \cite{Sundermeyer18,Li18, Material_Li18, Guo18}, 
but not without addressing the problem of bridging the domain gap between synthetic and real images \cite{Peng2018Syn2RealAN}.

Defocus blur is a well-modeled physical effect, and as such, straightforward to simulate in a realistic way.
The main insight of our work is that, while appearance features greatly differ from synthetic to real images, blur does not. Such domain invariant measurement effectively aids in bridging the domain gap between synthetic and real data.
We therefore propose to leverage defocus in a data-driven model to predict depth from focal stacks. By breaking the depth prediction into two steps, and using defocus maps as intermediate representations, we can train a neural network that generalizes from synthetic to real data without fine-tuning.
Additionally, we show our architecture works for an arbitrary number of input images and propose an extension to dynamic focal stacks \cite{VideoDFD_Kim16}, where camera motion or scene motion is present. 
Our {\bf contribution} is three-fold:
\begin{itemize}
    \item We propose to use defocus blur as intermediate supervision to train data-driven models to predict depth from focal stacks. We show that this is key towards generalization from synthetic to real images, and show state-of-the-art results.
    
    \item We generate a new synthetic dataset with multiple objects, textures and varying illumination with depth, blur and all-in-focus information.
    
    \item We propose architectures for static scenes that can work with a varying number of inputs. In dynamic focal stacks, our model can handle scene or camera motions within the stack. We show their robustness in a comprehensive ablation study.

\end{itemize}

\begin{figure}
\begin{center}
\resizebox{0.5\textwidth}{!}{
\includegraphics[width=1.0\textwidth, trim={0.0cm 4.25cm 3.7cm 0.0cm},clip]{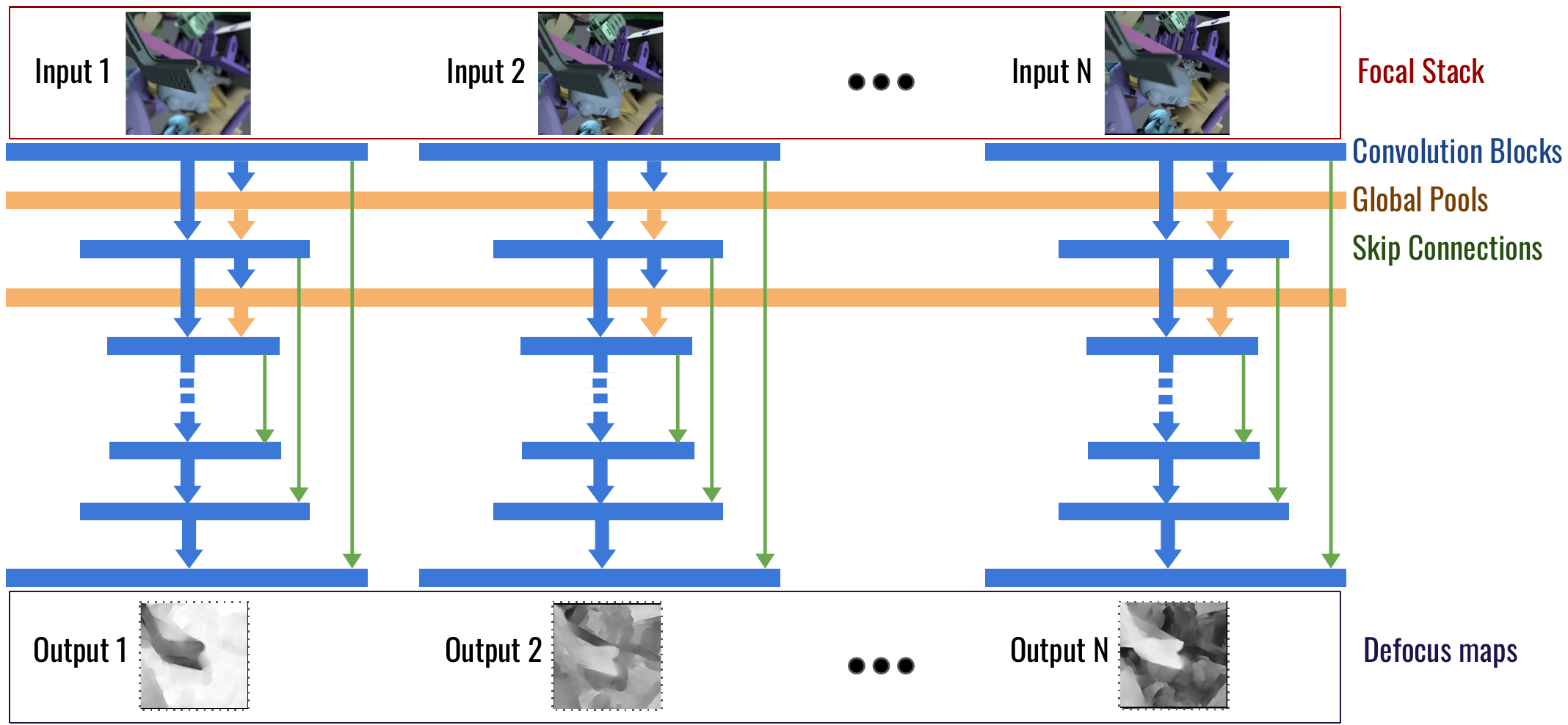}
}
\end{center}
   \caption{DefocusNet architecture. The proposed architecture takes a focal stack with an arbitrary size as an input and estimates corresponding defocus maps. The network uses an autoencoder as a basis and shares weights across all branches. Global pooling is used as a communication tool between separate branches.
   }
\label{fig:pipeline}
\end{figure}

\section{Related work}

\noindent{\bf Depth estimation from defocus.}
Depth estimation is a popular topic and is being explored from multiple directions. 
The vast majority of work focuses on monocular~\cite{Guo18, DBLP:conf/cvpr/FuGWBT18,Yin_2019_ICCV} or stereo \cite{DBLP:conf/cvpr/ZhangPYT19, DBLP:conf/cvpr/SmolyanskiyKB18} depth estimation. Using defocus information for depth prediction is less common.
Several optimization-based works~\cite{Suwajanakorn15,Surh17} estimate depth from a focal stack, while \cite{VideoDFD_Kim16} extends such methods to videos. These are general approaches that work on a variety of scenes, though they struggle on texture-less surfaces, and produce compelling depth measurements, but they are highly time-consuming and require careful calibration. It takes up to minutes for these methods to estimate the sharpness of the image regions and compose a depth image.
Recent methods leverage deep learning to bring this process closer to real-time. \cite{Hazirbas18} uses convolutional neural networks (CNNs) to estimate depth directly from input focal stacks, without considering the underlying image formation process. Such a method is bound to have generalization problems unless train and test conditions are very similar.  Additionally, it can only take a pre-defined number of inputs and does not incorporate any distance measurement.

Other works \cite{DepthEA_Anwar2017, Srinivasan18, Carvalho2018icip} implicitly use defocus information for monocular depth estimation. \cite{Srinivasan18} proposes to use defocus as a part of the loss function to estimate depth from an all-in-focus image. Nonetheless, they still use an all-in-focus monocular image as input, hence they do not leverage defocus blur during inference. 
Similarly, \cite{DBLP:conf/cvpr/GurW19} uses a differentiable loss layer that uses focus as a cue for depth prediction.
\cite{DepthEA_Anwar2017} uses CNNs for image deblurring and depth estimation from a single out-of-focus image. \cite{Carvalho2018icip} uses out-of-focus images for direct depth estimation. Their findings indicate that training on defocus images gives better depth measurements than training on sharp in-focus images. %
These methods use single images as input, therefore failing to leverage the much richer focus information present in a focal stack. As a consequence, they face difficulties when predicting depth in the wild, \ie, for completely different scenes and/or cameras than the ones used at training time.
In contrast, we propose to combine the power of data-driven approaches with knowledge of the image formation process, so that depth estimation can be computer by relying on focus differences between images in a focal stack.

\noindent{\bf Synthetic-to-real.}
There are several previous works that show domain generalization from synthetic to real data. According to \cite{Sundermeyer18}, all of them can be divided into three main strategies: domain adaptation, photo-realistic rendering, and domain randomization. 
Domain adaptation approaches usually convert samples from one domain to another \cite{DBLP:conf/cvpr/ShrivastavaPTSW17, DBLP:conf/cvpr/ZhaoFGT19}, or use model fine-tuning on real data after training on synthetic \cite{DBLP:conf/eccv/GuoLYRW18}. 
Several works show compelling results on photo-realistic synthetic training and real test sets \cite{Li18, Material_Li18}. However, photo-realism consumes a lot of time with physically based rendering (PBR) computation and hand-modelling of the entire environment.
Domain randomization, similar to data augmentation, introduces a variety of random attributes to make a model invariant to small changes and force the network to focus on the main features of the image. 
It requires simplistic modeling by assuming a random environment while being able to incorporate real data, \eg, in the background.
The main issue is to select the extent and type of attribute randomization. Several works show promising results in this direction \cite{Sundermeyer18, Aittala18}. 

We believe there is a fourth strategy for domain generalization, \ie, domain invariance, which involves training a model on features that are invariant for the two domains. 
One good example is stereo depth estimation. Models trained on synthetic data \cite{Guo18} are successful since they focus on the difference between two input images rather than the appearance and shape of objects.
This approach can also be embedded into a network architecture \cite{Aittala18}, where a permutation invariant architecture is proposed in an image translation context. It combines information across a random unordered number of input images. 

In our work, we rely on a domain invariant cue, defocus, as the main strategy for bridging the synthetic and real domain. Additionally, we also use domain randomization, and, to some extent, a photo-realistic rendering approach. %


\section{Learning depth through defocus}

In this section, we detail our model for depth estimation using defocus cues. We show that decomposing the problem into defocus estimation and later depth estimation is critical to close the domain gap between synthetic and real data. We show state-of-the-art results on real data while training our models only with synthetic data.

\subsection{Method overview} 
\label{sec:overview}

We show a diagram of our method in Fig.~\ref{fig:cover}, which shows the three main elements of our model:

\noindent{\bf DefocusNet.} Instead of directly estimating depth from a stack of RGB images, we first estimate a defocus map using DefocusNet (Section \ref{sec:defocus}). In particular, we estimate the amount of defocus per pixel.

\noindent{\bf DepthNet.} This defocus map is used as input to DepthNet, which estimates the scene depth map (Section \ref{sec:depth_est}).
To have sharper and better structured output depth, skip connections are used between the encoder of DefocusNet and the decoder of DepthNet.
Both DefocusNet and DepthNet are trained jointly and in an end-to-end manner.

\noindent{\bf AiFNet.} While obtaining a depth map is our end goal, image post-processing applications, \eg, refocusing, further require an all-in-focus (AiF) image, aside from the predicted depth map. 
We can easily predict the all-in-focus image with the focal stack and the defocus map, therefore, it is trivial to extend our architecture with a head, AiFNet, to predict all-in-focus images (Section \ref{sec:aif_est}).

In the following sections, we introduce the necessary concepts related to depth and defocus, and proceed to describe the three modules in detail.

\subsection{Defocus (blur) estimation} \label{sec:defocus}

\begin{figure}[t]
\begin{center}
 \includegraphics[width=0.48\textwidth, trim={0.0cm 0.75cm 0cm 0.0cm},clip]{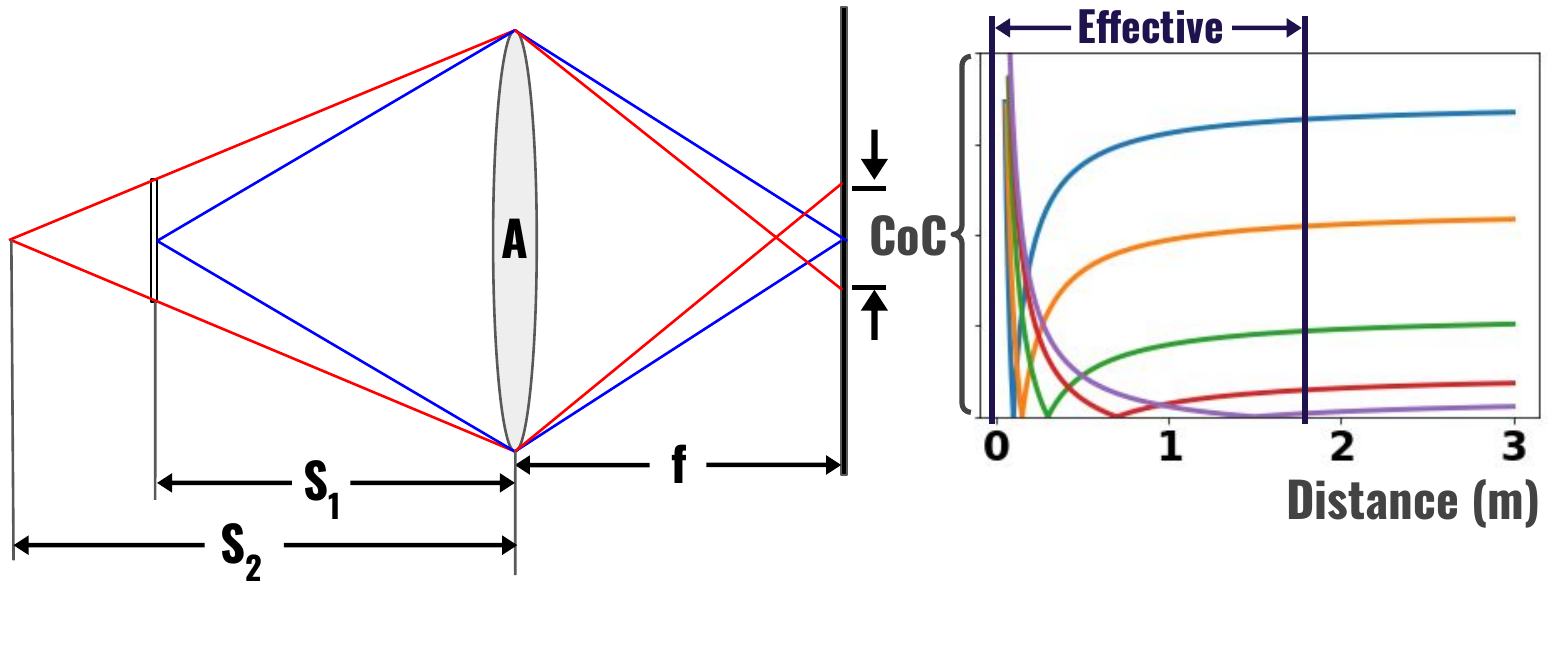}
\end{center}
   \caption{Lens diagram on the left. Circle of confusion plot on the right. Each line in the plot corresponds to a different focus distance. }
\label{fig:CoC}
\end{figure}

\noindent{\bf Circle of Confusion.}
In order to compute a defocus map, we first need to establish a measure for sharpness. %
When a light source passes through the camera lens, the light rays converge to form a focal point, which is found on the image plane of the camera. The circle of confusion (CoC) measures the diameter of the focal point. 
For a given point in front of the camera, if all rays of light flowing out of it converge into single location in the image plane, then the point will have sharpest projection possible (Fig. \ref{fig:CoC}).
Hence, the CoC is a direct translation of the amount of sharpness, equivalently, the amount of defocus.

The CoC can be computed using the following equation:
\begin{equation} \label{eq:coc}
c = \frac{\mid S_2-S_1 \mid}{S_2} \frac{f^2}{N(S_1-f)},
\end{equation}
where $f$ is the focal length of the lens, 
$S_1$ is the focus distance, $S_2$ is the distance from the lens to the object, and $N$ is the f-number. 
The f-number is the ratio of focal length to effective aperture diameter, essentially indicating aperture size. An illustration of a lens system is shown in Fig.\ref{fig:CoC}.  
The range of acceptable values of CoC, that we consider sharp, depends on the image format and camera model, and it is typically decided based on visual acuity. This range is referred as depth of field (DoF). 

On the right of Fig. \ref{fig:CoC}, we show a graph of the evolution of CoC values as the object distance $S_2$ increases. Each line represents a different value of focus distance $S_1$. 
The more variation lines produces in observation, the easier it is to estimate depth. 
Once the value surpasses the minimum diameter, the ambiguity of the CoC increases with the distance. 
After after a certain depth, the CoC no longer changes, indicating we can no longer rely on defocus cues to compute the depth of objects.
Thus, the defocus information is useful in a short range which depends on the camera properties. 

In this paper, we use the term defocus (blur) and CoC interchangeably. 
In fact, to construct a defocus map, we compute the CoC values for all pixels in an image, clip all values inside a chosen upper limit, and normalize all values in a range between 0 to 1 (from sharp to blurry). 

\noindent{\bf Depth-from-Defocus limitations.}
The problem is inherently limited by design, as depth-from-focus works better on short ranges.
Nonetheless, ubiquitous depth-from-stereo methods (and by extension, conventional depth cameras with a separate IR projector and an IR camera) are less effective in a short range due to part of the scene being not visible by both cameras. 
Depth-from-focus can be seen as a solution for short ranges. 
In our camera settings, the effective range in which we can use defocus to predict depth is within 2 meters, as shown in Fig.~\ref{fig:CoC}.

\noindent{\bf Data-driven defocus estimation.}
A key design choice of our work is to use defocus estimation as an intermediate step, or supervisory signal, to estimate depth from a focal stack. This allows us to obtain a model that generalizes from synthetic to real images.  
Therefore, we begin by training a model, DefocusNet, to estimate a defocus map from a set of images.
Fig.~\ref{sec:dataset} shows an example of RGB images from our synthetic dataset and their corresponding defocus map, where the pixel values represent their sharpness level.

The focal stack is processed by an autoencoder convolutional neural network (CNN), as shown in Fig.~\ref{fig:pipeline}.
The encoder contains one branch per image in the focal stack, where all branches share the weights of the CNN. 
Our goal is to encourage the network to perform comparisons between the features extracted at each branch, as to better establish focused and defocused regions. 
We do this comparison at every layer of the CNN, inspired by \cite{Aittala18}.

\noindent{\bf Layer-wise global pooling.} The network computes the output of the convolution layer for every input image, then all output feature maps are pooled by a symmetric operation, \ie, we compute the maximum value of each feature map cell across all branches.
The globally pooled features, or {\it global} feature map, are then concatenated to the {\it local} feature map coming from each branch. The combined output is then passed to the next convolution layer, and the process is repeated. 
This way, each CNN branch will contain both local as well as global features. 
Intuitively, this allows the network to compare local features with globally pooled features, finding out the sharpest regions of the image by comparison, and passing those to the next layer of the CNN.
The main advantage of using this layer-wise pooling across inputs is that our model can handle an arbitrary number of images as input, making our model extremely flexible.

The CNN is rather shallow with only 4 layers. Since sharpness is a local property, we do not need a large receptive field.  
The decoder then propagates the estimated focus information from the edges to the center of the objects, where there might not be enough texture to properly estimate sharpness. We also make use of skip connections (by concatenation) to properly recover boundaries in the regressed defocus maps.
Global pooling is used only in the encoder, while the decoder has separate branches for each output. The main idea is for the encoder to learn to detect sharp regions by comparison, and for the decoder to regress a defocus map independently for each input.
We use an L2 loss to train DefocusNet without additional regularization.

As we can see in Fig. \ref{fig:cover}, once defocus is estimated, we can use it as an input to estimate depth. 

\subsection{Depth estimation.} \label{sec:depth_est}
A depth map can be constructed from defocus maps by using the camera capture settings for each image in the focal stack.
However, our network architecture is by design unaware of image order. 
Therefore, we also include a focus distance map together with the previously predicted defocus map as input to DepthNet, our neural network that is trained for depth regression. The focus distance map is a single channel and single value image,
where every pixel takes the value of the focus distance.

We obtain focus distances from our dataset rendering script (Section \ref{sec:dataset}) and then rescale them in the range between 0 and 1.
For real images, if we know the order in a focal stack, we can assign focus distances consecutively with computed increments (based on a number of images) in the range 0 to 1. We can also extract the focus distance from EXIF properties (camera settings used to take an image) and rescale the values to the required range. 
While our architecture needs this additional input, it comes at a minimal cost, does not affect generalization and allows us to have an arbitrary number of input images.

The network architecture for DepthNet is similar to DefocusNet, except we have a single branch also in the decoder to get one depth map as output.
Additionally, in the DepthNet decoder we use
skip connections from the DefocusNet encoder to combine information from the RGB input image. This helps especially to improve the depth prediction around object boundaries.
During training, we use the L2 loss between estimated and ground-truth depth.

The full loss function is shown below:
\begin{equation}
\label{eq:depth_loss}
\mathcal{L} = \lambda_{a}\sum \| I_{def} - E_{def}\|_{2} + 
  \lambda_{b}\sum \| I_{dep} - E_{dep}\|_{2}
\end{equation}

where $I$ are ground truth images, $E$ are estimated images, for depth and defocus, $\lambda_{a,b}$ are weight coefficients.

\noindent{\bf Dynamic stacks.}
As we mentioned in previous paragraphs, the goal of global pooling across inputs is to encourage the network to compare different input branches. 
However, such approach has a problem with focal stacks taken with a moving camera or a static camera but a moving scene. In those scenarios, each part of the image will contain drastically different information, and comparison across inputs will not be informative to compute defocus.
To handle such scenarios, we propose to use a recurrent autoencoder \cite{DBLP:journals/tog/ChaitanyaKSSLNA17} for DepthNet and DefocusNet, instead of the global pooling autoencoder.
Such a recurrent autoencoder concatenates the local features from one branch to the next sequentially, taking order into account. 
This allows us to gradually incorporate changes, also changes in the scene, and implicitly compare the amount of defocus in them.
Still, such recurrent architecture has its own drawbacks: (i) it has a short memory \cite{DBLP:journals/tnn/BengioSF94}, and (ii) the number and order of images on the focal stack is fixed due to the architecture. 
For these reasons, we use it only when dealing with dynamic stacks.
More architecture details are given in the supplementary material.

\subsection{All-in-Focus estimation} \label{sec:aif_est}

In our work, we use a stack of differently focused images to estimate a depth map. However, for image post-processing applications such as refocusing, we additionally need an all-in-focus (AiF) image where all pixels are appropriately sharp.
We can estimate such image given a focal stack by combining different image parts according to their sharpness. 
For this reason, we propose to incorporate such estimation inside our network, reusing the focal stack processing as well as the defocus map. The AiF image is computed by an additional CNN head, the AiFNet.
Since AiF prediction is not the main focus of our work, the model description and results are included in the supplementary.

\section{Synthetic Training Data} \label{sec:dataset}
\begin{figure}
\begin{center}
\resizebox{0.47\textwidth}{!}{%
\includegraphics[width=\textwidth, trim={0.0cm 0cm 0cm 0cm},clip]{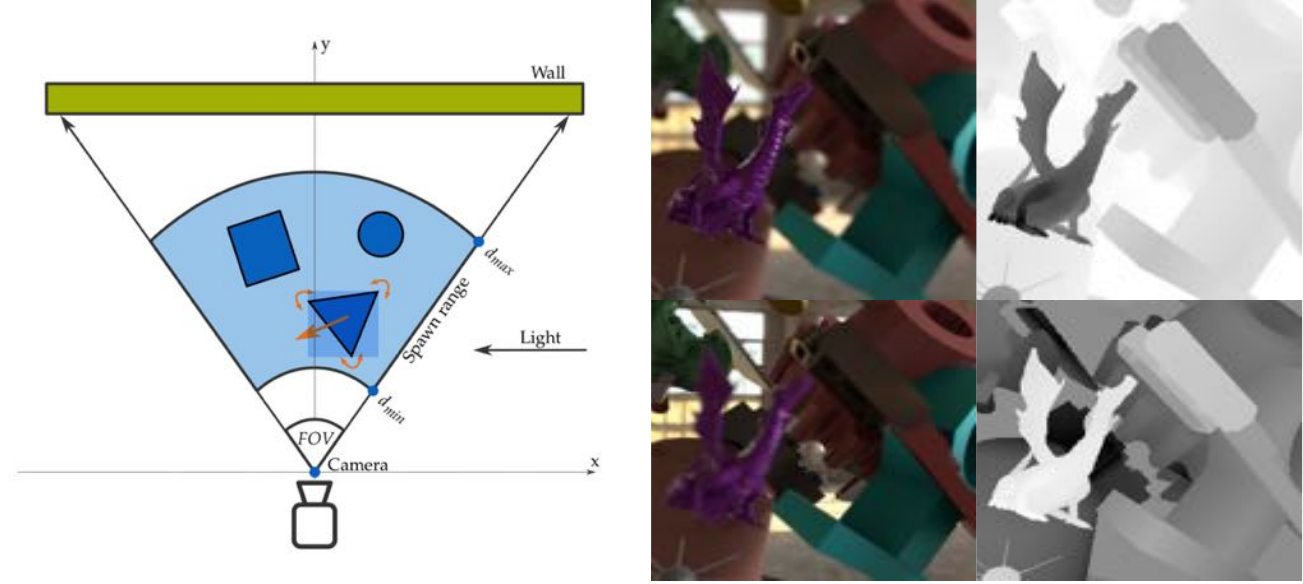}
}
\end{center}

   \caption{On the left is a schematic for the random scene generation. On the right side are the examples from the synthetic dataset (pairs of RGB images and defocus maps from focal stacks). Sharper regions are darker in defocus maps.
   }
\label{fig:SynthExample}
\end{figure}

We use only synthetic data to train our full model for depth prediction from focal stacks. As we will show in the experimental section, our network will generalize to depth prediction on real images without the need to fine-tune our model on real data.
We create our synthetic dataset using Blender \cite{blender2018} Cycles renderer with reflection turned on, but without shadows. Examples from the dataset are shown in Fig.~\ref{fig:SynthExample}.
For training, we render a total of 1000 scenes, each of them with 5 RGB images per focal stack, 5 defocus maps, 1 depth image and 1 all-in-focus image (taken with a wide aperture).
Each defocus map was calculated using Equation~\eqref{eq:coc} based on depth and camera parameters. Each image was rendered at a resolution of $256 \times 256$.

\noindent{\bf Dataset characteristics.} We acquired CAD 3D objects from a model repository containing a total of 400 objects \cite{Thingi10K}, and place between 20 to 30 objects per scene, all randomly chosen. Objects are assigned a random size, location and rotation in each scene to have a random spatial arrangement. Locations are limited to the effective range of defocus (Sec. \ref{sec:defocus})  and camera field of view. 
Some objects might not be fully in the camera field of view or can be occluded by other objects.
We choose a non-realistic scene composition on purpose due to its simplicity to model, but primarily to avoid overfitting on spatial cues, and instead force the model to focus on defocus cues.

Each object is assigned one random material. Before
rendering, we randomize the hue of the diffuse and specular components, glossiness and roughness are chosen within a range that produces a realistic appearance. All materials use physically based rendering (PBR) shaders.
For illumination, we used 20 different HDR environment maps (EM), both indoor and outdoor.
We used fixed camera parameters with fixed focus distances for all scenes. The f-number is set to 1 in order to have a shallow depth-of-field, therefore making depth changes more observable in terms of blur.

\noindent\textbf{Dynamic stacks.} To handle camera or scene motion, we additionally create dynamic stacks with 4 images, where the position of the object changes for each image. We assign a random direction and magnitude of translation and rotation for each object at the beginning of sequence rendering. More details are provided in the supplementary material.


\vspace{-1em}
\section{Evaluation} \label{sec:evaluation}
We present a comprehensive ablation study on a diverse set of synthetic scenes. We further show qualitative and quantitative results for depth prediction on real datasets and also on real images with synthetic blur. We show generalization by training only on synthetic data and obtaining state-of-the-art results on real images with synthetic blur.
We provide in the supplementary material additional results on real/synthetic images and all-in-focus image prediction.

\vspace{-1em}
\subsection{Implementation details}
The method is implemented in PyTorch\cite{pytorch}. We train the model using the
Adam optimization algorithm \cite{adam_opt} with a learning rate of 0.0001.
We assign $\lambda_{d}=0.02$ and $\lambda_{a,b}=1$. 
We provide a detailed description of all network architectures in the supplementary material.

\noindent\textbf{Run-Time Performance.} On an Nvidia Titan X, a forward pass of our network takes 70ms on a focal stack with 10 images and 150ms with 20 images. For \cite{Suwajanakorn15} reported time is 20mins and for \cite{Surh17} - 6.7s.

\subsection{Metrics and Datasets} \label{sec:test_dataset}

\noindent\textbf{Evaluation metrics.} For depth comparison, we use root mean squared error (RMSE) in Table \ref{tab:depth_real2} and mean squared error (MSE) in all other tables in order to compare to existing methods.\\

\vspace{-0.5cm}
\noindent\textbf{Synthetic dataset.}
Synthetic test data was rendered in the same way as the training data in Section \ref{sec:dataset}, but we use a new set of 10 environment maps, 20 new textures, and 300 new objects. 
We rendered 4 test sets to evaluate generalization:

\noindent{\it (i) Shape test.} The first test set contains only new objects, while environments and textures are the same as used for training. This is a test on shape generalization.

\noindent{\it (ii) Appearance test.} The second test set contains new objects, textures and environment maps to test appearance generalization.

\noindent{\it (iii)  Wide DoF test.} The third test is identical to the second set, but with smaller camera apertures (f-number is from 3 to 10)  to show that we rely on defocus cue. Having a larger DoF negatively affects blur estimation because there is less blur information. 

\noindent{\it (iv) Medium DoF test.} The fourth test is rendered with the same options as the second set, but with  slightly smaller camera apertures (we randomly choose an f-number from 1 to 3 whereas a training set was trained on f-number of 1) to test the generalization to slightly different camera settings.

\noindent\textbf{Real datasets.} There are very few datasets that provide focal stacks. We use these datasets for quantitative and qualitative experiments.

\noindent{\it (i) DDFF 12-Scene benchmark~\cite{Hazirbas18}.} Obtaining focal stacks with standard cameras is a time-consuming task. To speed up the process, \cite{Hazirbas18} uses plenoptic cameras that can capture 4D light-fields. With a single light-field, we can generate a focal stack and an all-in-focus image. 
This dataset consists of 1200 focal stacks with 10 images each. 
The dataset is challenging due to the type of scene recorded: many flat and texture-less surfaces such as walls and desks, and other texture-less objects such as monitors, doors and cabinets. Furthermore, their capture settings are not optimal for defocus blur, they shoot with wide DoF and capture scenes with far distances.
We use the same training/test split as~\cite{Hazirbas18}. 

\noindent{\it (ii) Mobile Depth~\cite{Suwajanakorn15}.} This dataset consists of 13 scenes, each scene has a different number of images in the range between 13 and 32. All images were taken with a mobile phone and aligned using optical flow.
Since there are not enough images to train a deep learning approach, we show the results of our model trained only on our synthetic training set.\\

\vspace{-0.5cm}
\noindent\textbf{Synthetically blurred real datasets}. Due to lack of datasets with focal stacks for quantitative experiments, we propose to use popular indoor datasets and create focal stacks from RGB images and ground-truth depth. We apply synthetic blur following~\cite{DBLP:conf/cvpr/GurW19}.

\noindent{\it (i) NYU Depth Dataset v2~\cite{Silberman:ECCV12}}. 
This dataset consists of 1449 pairs of aligned RGB and depth frames. We use the regular split between test and train sets.

\noindent{\it (ii) 7-Scenes~\cite{DBLP:conf/cvpr/ShottonGZICF13}}. This dataset consists of around 43000 images of aligned RGB and depth frames scenes. Due to large size of the dataset, we randomly sample total of 890 images from all 7 sequences for our tests.

\noindent{\it  (iii) Middlebury stereo dataset~\cite{DBLP:conf/dagm/ScharsteinHKKNWW14}}. This dataset consists of 46 images of stereo RGB pairs and disparity frames. Based on provided camera calibration parameters, we compute the depth map for the RGB image corresponding to the right camera. 

\noindent{\it (iv) SUN RGB-D~\cite{DBLP:conf/nips/ZhouLXTO14}}. This dataset is a combination of NYU depth v2\cite{Silberman:ECCV12}, Berkeley B3DO\cite{DBLP:conf/iccvw/JanochKJBFSD11} and SUN3D\cite{DBLP:conf/iccv/XiaoOT13} datasets with improved depth maps. It consists of 10,000 images of aligned RGB and depth frames. Similarly, we randomly sample total of 490 images for our tests.

\input{tables/depth_table2}
\input{tables/depthone_table}

\subsection{Ablation Study on Synthetic Dataset}
We quantitatively analyze our method’s generalization capabilities to new environments and camera settings on 4 different synthetic test modalities.
Table \ref{table:SynthDepth} shows estimation results between different models on the depth estimation task.  
We explain and compare all models below.
In the tables, "All" indicates that all $N$ images of the focal stack were used for prediction, while "Random" indicates the use of a randomly chosen number of images $r \leq N$.

\noindent\textbf{Architecture choice.} We consider two architecture types: (i) FixedAE, our baseline with single autoencoder without global pooling layers that is trained on a single input with fixed-sized focal stack with $N$ images, and (ii) PoolAE, the autoencoder with global pooling layers, which can take any number of images as input. 
In Table \ref{table:SynthDepth}, the first two rows show a direct comparison of the two models. 
FixedAE was trained on a specific number of inputs and shows good performance only when tested on the same number of images (column All).
In contrast, PoolAE shows generalization to the column Random, where the number of input images varies but accuracy is maintained.
From these tests, we can summarize that PoolAE architecture is robust, generalizes better than just stacking all inputs in single AE, and has the added value of processing any number of inputs without the need to retrain the model.

\noindent\textbf{Is defocus needed?} We compare models that directly estimate depth from the input RGB image and our proposed approach, where we first estimate a defocus map and then depth.
In Table \ref{table:SynthDepth}, with our PoolAE architecture, we achieve a much better result when going over the defocus map (row 3.) compared to a direct estimation (row 2.). 
On the {\it Wide DoF test}, PoolAE with defocus map (row 3.) performs worse due to having less defocus blur to rely on. It confirms that our model uses defocus as the primary signal to estimate depth. If that signal is inexistent due to a too wide DoF, performance drops as expected.
Overall, we can clearly conclude that using defocus information is beneficial for depth estimation, and is cue that allows us to generalize to a diverse set of object shapes and appearances.

\begin{figure}
\begin{center}
\resizebox{0.4\textwidth}{!}{%
\includegraphics[width=\textwidth, trim={0.0cm 0cm 0cm 0cm},clip]{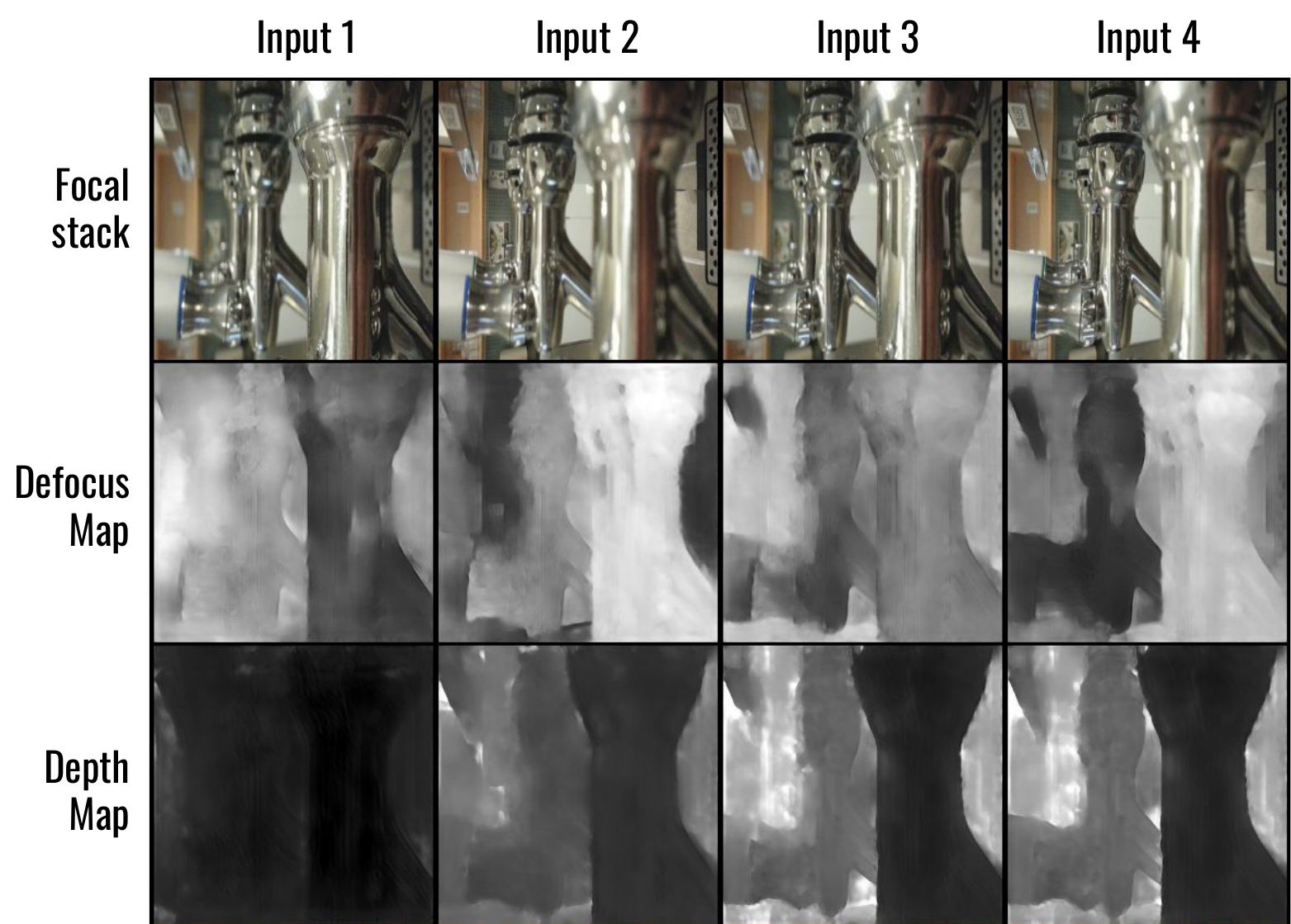}
}
\end{center}
\vspace{-0.3cm}
   \caption{Our sequential estimates on a real focal stack. The first row images are inputs to our pipeline. The other horizontal sequences show our outputs for a growing number of input frames. So the results on the left use only the first input image and on the right use all four inputs. Note how adding more inputs quickly improves the depth.
   }
\label{fig:SynthDefocus}
\end{figure}

\begin{figure}
\begin{center}
\includegraphics[width=0.5\textwidth, trim={0.0cm 0.cm 0cm 0.0cm},clip]{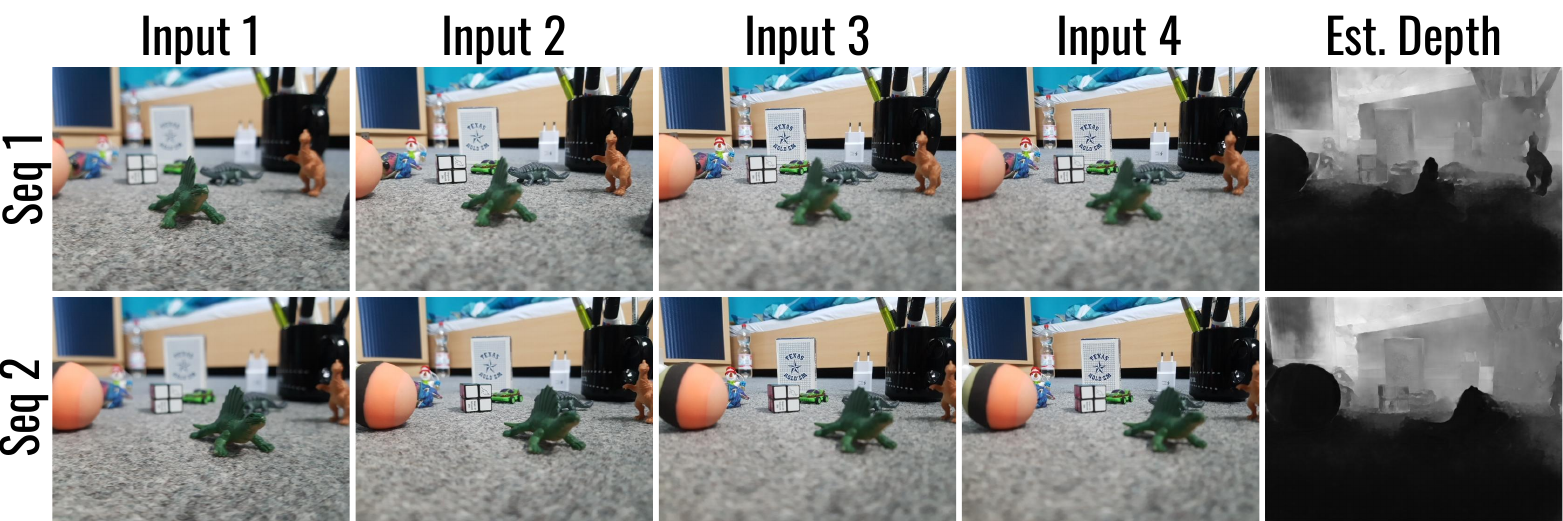}
\end{center}
\vspace{-0.3cm}
   \caption{Our estimates on a real focal stack on a dynamic sequence.}
\label{fig:VideoSeq}
\end{figure}

\input{tables/depth_real5}

\begin{figure}[t]
\begin{center}
   \resizebox{0.5\textwidth}{!}{%
\includegraphics[width=\textwidth, trim={0.0cm 5.6cm 0cm 0.1cm},clip]{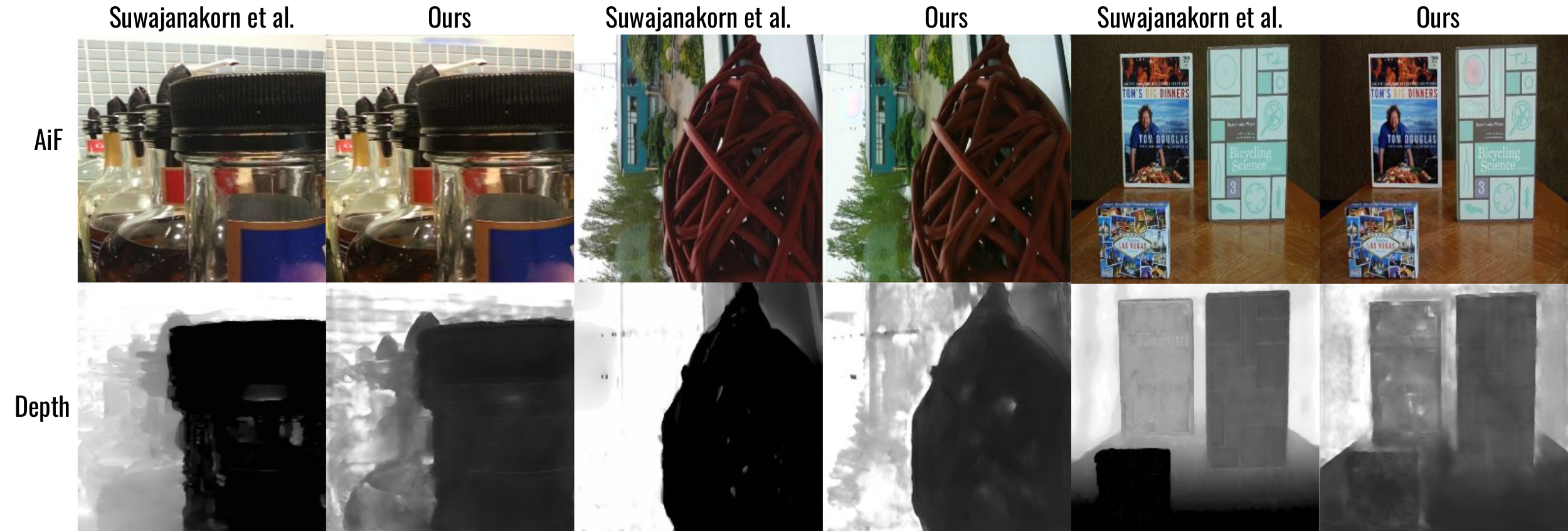}
}
\end{center}
\vspace{-0.3cm}
   \caption{Qualitative results on Mobile Depth dataset. We compare our method with a model-based approach. \cite{Suwajanakorn15}.}
\label{fig:RealDepthMobile}
\end{figure}

\noindent\textbf{Single image vs. Focal stack.}
We also train our method on a single image input setting.
Such network is expected to perform worse when predicting depth, as it is not able to rely on blur comparison between images in the focal stack.
Results are presented on the first row of Table \ref{table:SynthDptOne}. We clearly see the error in depth prediction is much larger than all  models that use a focal stack (Table \ref{table:SynthDepth}).

\noindent\textbf{All-in-Focus vs. Out-of-Focus.}
Out-of-focus images give more information related to depth, as shown in \cite{Carvalho2018icip}. We also perform a similar test in Table \ref{table:SynthDptOne}, where we compare to a model trained on all-in-focus images. 
We can see from the {\it Shape} and {\it Appearance} tests, that defocus gives more information for depth even without explicitly computing the defocus map. 
The model in Row 1. also performs worse in both camera aperture tests, since they both have wider DoF. Wider DoF has less defocus blur and does not give any information to the model. 
The Row 2 model does not rely on defocus and therefore shows similar results on all tests.

\input{tables/ddff_table2}
\input{tables/depthmobile_table2}

\vspace{-1em}
\subsection{Evaluation on Real data}

\noindent\textbf{Synthetically blurred NYU, 7 scenes, Middlebury and SUN RGB-D.}
In this section, we show quantitative results on cross-dataset and domain generalization task. 
Since defocus blur is only effective on close distances, we conduct experiments on 4 different versions of each dataset: (i) {\it regular}, no modifications, (ii) {\it less than 2m}, only depths within 2 meters are taken into account, since this is the functioning range for our method, (iii) {\it normalized version}, depth is rescaled from 0 to 10 meters to a range from 0 to 1 meters, and (iv) {\it 45 degrees}, the normalized version with images rotated 45 degrees. 
The last test of rotating input images by 45 degrees is a simple yet effective way to show that current datasets have photographic bias, which leads to overfitting on the training settings.

We compare our method with the state-of-the art single image depth estimation model VNL \cite{Yin_2019_ICCV} in Table \ref{tab:depth_real2}.
We trained our method on the regular synthetic dataset, "Ours", and its normalized version, "Ours*". We additionally show a version fine-tuned on the NYU dataset.
 Due to the difference in datasets, we use the median to rescale estimated depth for all models to match ground-truth depth as in \cite{Gordon19}.
VNL was trained on the NYU dataset and performs well on that dataset. However, on other datasets, its performance drops in comparison to our methods that use defocus cues.
Our fine-tuned versions, Ours(Synth.+NYU) and Ours*(Synth.+NYU), perform best across all test but fail the 45 degree test.  
The purely synthetically-trained model shows similar or better performance to the method trained purely on real data, and generalizes much better in the case of the 45 degree experiment.
This clearly shows the bias of the real dataset that does not allow the networks to generalize to any scene configuration. 

We use focal stacks with 4 images which is to some degree unfair to single image methods. Nonetheless, capturing a focal stack is straightforward: (i) it takes just slightly longer than a single shot, (ii) we do not need additional camera hardware, and (iii) we do not need to move the camera to satisfy stereo requirements. 
The benefits in depth prediction accuracy come at a very little cost during caption.

\noindent\textbf{DDFF 12-Scene.} We compare our approach 
to a CNN-based method~\cite{Hazirbas18} and a classic method (VDFF) \cite{Moeller2015}.
As explained in Section \ref{sec:dataset}, this dataset is not ideal for DefocusNet due its wide DoF. 
Our synthetically trained models did not directly perform well, but after fine-tuning on the provided training data, we were able to achieve state-of-the-art results, shown in Table \ref{table:RealDDFF}. 
As we can see, PoolAE network with DefocusNet shows better performance. We conclude that our approach generalizes well within similar types of real images and is able to handle texture-less surfaces.

\noindent\textbf{Mobile Depth from Focus.}
We compare our method with traditional methods \cite{Suwajanakorn15} that take focal stack images as inputs. The dataset does not have ground truth depth, but the authors provide their depth estimations. We compare our models to their depth to test the generalization capabilities of our approach. Table \ref{table:RealMobileDepth} shows that a direct approach does not generalize from synthetic to real images while our method does.
We also show qualitative results in Fig.~\ref{fig:RealDepthMobile} and Fig.~\ref{fig:SynthDefocus}. Note that our models are {\it trained on synthetic data only}, and are not fine-tuned on this dataset. 
Additionally, we show visual results with the increasing number of input images in Fig.~\ref{fig:SynthDefocus}. We can see that it gradually improves the depth estimates thanks to our pooling architecture.

There are several aspects that work in favor of a better generalization in our work: 
(i) we use down-scaled images, e.g., original images from Mobile Depth are 360 x 640, which makes the out-of-focus blur details similar for most conventional cameras; (ii) the method is based on the comparison between differently focused inputs rather than analysis of blur shape/size; (iii) \cite{Aittala18} showed that with enough randomness in synthetic noise, invariance to various real noise can be achieved.

\noindent\textbf{Dynamic stacks.} Since we lack real test data to show our model on dynamic stacks, we implemented a smartphone application to capture focal stacks. Fig.~\ref{fig:VideoSeq} shows qualitative results of the recurrent approach on real data recorded with a moving camera. Note, the models were trained with synthetically moving sequences.
We show more qualitative results for all datasets in our supplementary material.

\section{Conclusion}

We presented a data-driven approach for estimating depth using defocus cues from a focal stack as a supervisory signal. 
Our key design decision is to use domain invariant defocus information as supervision for the depth prediction. This allows our model to generalize from synthetic to real images.
Our permutation-invariant network allows us to correctly estimate depth with any focal stack size, and we further show a simple extension to process stacks with either moving camera or moving scene.

\textbf{Acknowledgements.} This research was funded by the Sofja Kovalevskaja Award of the Humboldt Foundation.

{\small
\bibliographystyle{ieee_fullname}
\bibliography{bib}
}


\end{document}

%% file: tables/depth_table2.tex
\begin{table*}[]
\begin{center}
\resizebox{1.\textwidth}{!}{%
\begin{tabular}{l|c|c||c|c||c|c||cc}
\hline
\multicolumn{1}{c|}{}                                  & \multicolumn{2}{c||}{\it Shape}                                 & \multicolumn{2}{c||}{\it Appearance}                                 & \multicolumn{2}{c||}{\it Wide DoF}                                 & \multicolumn{2}{c}{\it Medium DoF}                                                       \\
\multicolumn{1}{c|}{\multirow{-2}{*}{Models}}          & All                          & Random                       & All                          & Random                       & All                          & Random                       & \multicolumn{1}{c|}{All}                          & Random                       \\ \hline
1. FS $\,\to\,$ Depth (FixedAE)                             & 0.014                        & 0.097                        & 0.012                        & 0.095                        & 0.103                        & 0.111                        & \multicolumn{1}{c|}{0.083}                        & 0.112                        \\
2. FS $\,\to\,$ Depth (PoolAE)                               & 0.031                        & 0.036                        & 0.034                        & 0.039                        & 0.049                        & \textbf{0.050}               & \multicolumn{1}{c|}{0.047}                        & 0.048                        \\

3. FS $\,\to\,$ Defocus $\,\to\,$ Depth (PoolAE)                    & \textbf{0.008}                        & \textbf{0.013}                        & \textbf{0.007}                        & \textbf{0.012}                        & \textbf{0.042}                        & 0.078                        & \multicolumn{1}{c|}{\textbf{0.014}}                        & \textbf{0.030}                        \\
\hline

\end{tabular}}
\end{center}
\vspace{-0.2cm}
\caption{Results on the synthetic data test sets for depth estimation models with focal stacks (FS) as input. {\it Row 1.} Direct depth prediction with a fixed-sized AE.  {\it Row 2.} Direct depth prediction with the proposed AE with global pooling. 
{\it Row 3.} Depth prediction for our method, using the predicted defocus map as supervisory signal.  }
\label{table:SynthDepth}
\end{table*}

%% file: tables/depthone_table.tex
\begin{table}[]
\begin{center}
\begin{tabular}{l|cccc}
\hline
\multicolumn{1}{c|}{Models} & {\it Shape}      & {\it App.}       &  {\it W. DoF}          &  {\it M. DoF}        \\ \hline
1. RGB $\,\to\,$ Depth             & \textbf{0.032} & \textbf{0.031} & 0.134          & 0.108          \\ \hline
2. AiF $\,\to\,$ Depth             & 0.049          & 0.050          & \textbf{0.050} & \textbf{0.054} \\ \hline
\end{tabular}
\end{center}
\vspace{-0.2cm}
\caption{Results of depth estimation on the synthetic test set using only one image as input, either one of the out-of-focus images of the focal stack (Row 1), or the all-in-focus image (Row 2).}
\label{table:SynthDptOne}
\end{table}

%% file: tables/depth_real5.tex
\begin{table*}[]
\resizebox{\linewidth}{!}{
\begin{tabular}{l|cc|cccc|cccc|cccc|cccc}
\hline
\multicolumn{1}{c|}{\multirow{2}{*}{Models}} &
\multicolumn{2}{c|}{Training data} &
\multicolumn{4}{c|}{NYU}                                          & \multicolumn{4}{c|}{7 scenes}                                     & \multicolumn{4}{c|}{Middlebury}                                   & \multicolumn{4}{c}{SUN RGB-D}                                    \\ \cline{2-19} 
\multicolumn{1}{c|}{}   & Synth. & NYU                      & Norm.*         & 45deg.*        & Regular        & \textless{}2m  & Norm.*         & 45deg.*        & Regular        & \textless{}2m  & Norm.*         & 45deg.*        & Regular        & \textless{}2m  & Norm.*         & 45deg.*        & Regular        & \textless{}2m  \\ \hline
Ours    &  \checkmark  &     & -              & -              & 1.054          & 0.272          & -              & -              & 0.504          & 0.282          & -              & -              & 0.803          & 0.384          & -              & -              & 0.721          & 0.259          \\
Ours*    & \checkmark  &                                & 0.056          & \textbf{0.073} & -              & -              & 0.030          & \textbf{0.037} & -              & -              & 0.052          & \textbf{0.063} & -              & -              & 0.037          & \textbf{0.052} & -              & -              \\ \hline
Ours    & \checkmark  &  \checkmark                         & -              & -              & 0.493          & \textbf{0.181} & -              & -              & \textbf{0.277} & \textbf{0.189} & -              & -              & \textbf{0.544} & \textbf{0.351} & -              & -              & \textbf{0.360} & \textbf{0.196} \\
Ours*    & \checkmark  & \checkmark                         & \textbf{0.013} & 0.111          & -              & -              & \textbf{0.010} & 0.045          & -              & -              & \textbf{0.025} & 0.079          & -              & -              & \textbf{0.014} & 0.073          & -              & -              \\ \hline
VNL \cite{Yin_2019_ICCV}    &   &  \checkmark                                 & 0.040          & 0.100          & \textbf{0.395} & 0.206          & 0.033          & 0.050          & 0.328          & 0.244          & 0.064          & 0.071          & 0.645          & 0.400          & 0.037          & 0.068          & 0.370          & 0.289          \\ \hline
\end{tabular}
}
\caption{Regular  - no modifications, \textless{}2m - same as regular but counting results only for depth less than 2 meters, and normalized version - depth was rescaled to range from 0 to 1. All models with * were trained for normalized sets.  45 degrees set is a version with images rotated 45 degrees. Our models trained first on synthetic dataset then tested with and without finetuning on NYU dataset. All tests show RMSE values.}
\label{tab:depth_real2}
\end{table*}

%% file: tables/ddff_table2.tex
\begin{table}[]
\begin{center}
\begin{tabular}{l|c}
\hline
\multicolumn{1}{c|}{Models}      & MSE                 \\ \hline
FS $\,\to\,$ Depth (F)          & 11.7 * 10\textsuperscript{-4}          \\
FS $\,\to\,$ Depth (P)            & 13.2 * 10\textsuperscript{-4} \\
FS $\,\to\,$ Defocus $\,\to\,$ Depth (P) & \textbf{9.1 * 10\textsuperscript{-4}}          \\
DDFF \cite{Hazirbas18} & 9.7 * 10\textsuperscript{-4}    \\
VDFF \cite{Moeller2015} & 73.0 * 10\textsuperscript{-4}            \\\hline
\end{tabular}
\end{center}
\vspace{-0.2cm}
\caption{Results of depth estimation on DDFF-12.}
\label{table:RealDDFF}
\end{table}

%% file: tables/depthmobile_table2.tex
\begin{table}[]
\begin{center}
\begin{tabular}{l|c}
\hline
\multicolumn{1}{c|}{Models}    & MSE         \\ \hline
FS $\,\to\,$ Depth            & 0.184          \\
FS $\,\to\,$ Defocus $\,\to\,$ Depth & \textbf{0.045} \\ \hline
\end{tabular}
\end{center}
\vspace{-0.2cm}
\caption{Results of depth estimation on Mobile Depth dataset.}
\label{table:RealMobileDepth}
\end{table}